\documentclass[runningheads]{llncs}
\usepackage[T1]{fontenc}
\usepackage{graphicx}
\usepackage{booktabs}
\usepackage{hyperref}
\usepackage{xurl}
\usepackage{xcolor}
\usepackage{graphicx}
\usepackage{svg}
\usepackage{subcaption}
\usepackage{amsmath}

\newcommand{\framework}{\textsc{AgentHeLLM~}}

\newcommand{\tool}{\textsc{AgentHeLLM Pathfinder~}}

\begin{document}

\title{Agent2Agent Threats in Safety-Critical LLM Assistants: A Human-Centric Taxonomy }

\titlerunning{Threat Modeling for In-Vehicle AI Agents}

\author{Lukas Stappen\inst{1} \and Ahmet Erkan Turan\inst{1} \and 
Johann Hagerer\inst{2} \and Georg Groh\inst{2}}

 \authorrunning{Stappen et al.}

\institute{BMW Group Research, Munich, Germany \and
 Technical University Munich, Munich, Germany}

\maketitle

\begin{abstract}
The integration of Large Language Model (LLM)-based conversational agents into vehicles creates novel security challenges at the intersection of agentic AI, automotive safety, and inter-agent communication. As these intelligent assistants coordinate with external services via protocols such as Google's Agent-to-Agent (A2A), they establish attack surfaces where manipulations can propagate through natural language payloads, potentially causing severe consequences ranging from driver distraction to unauthorized vehicle control. Existing AI security frameworks, while foundational, lack the rigorous ``separation of concerns'' standard in safety-critical systems engineering by co-mingling the concepts of \emph{what} is being protected (assets) with \emph{how} it is attacked (attack paths). This paper addresses this methodological gap by proposing a threat modeling framework called \textsc{AgentHeLLM}
(\emph{Agent Hazard Exploration for LLM Assistants})
that formally separates asset identification from attack path analysis. We introduce a human-centric asset taxonomy derived from harm-oriented ``victim modeling'' and inspired by the Universal Declaration of Human Rights, and a formal graph-based model that distinguishes \emph{poison paths} (malicious data propagation) from \emph{trigger paths} (activation actions). We demonstrate the framework's practical applicability through an open-source attack path suggestion tool \textsc{AgentHeLLM Attack Path Generator} that automates multi-stage threat discovery using a bi-level search strategy.

\end{abstract}

\section{Introduction}
\label{sec:introduction}

The automotive industry is undergoing a transformation marked by the integration of Large Language Model (LLM)-based conversational agents into vehicles~\cite{stappen2023integrating}. The latest voice assistant systems such as BMW Intelligent Personal Assistant (powered by Amazon Alexas LLM)~\cite{bmwGenerativeAI}, Volkswagen's IDA (powered by ChatGPT)~\cite{cerenceVWLaunch} and Mercedes-Benz's MBUX Virtual Assistant (powered by Google Gemini)~\cite{mercedesGemini2025} represent a shift from rigid command-and-control interfaces to intelligent, autonomous companions capable of natural conversation, tool use, and learning driver preferences over time.
These systems increasingly exhibit \emph{agentic} behavior: they run reasoning loops~\cite{yao2023react}, invoke tools, maintain persistent memory~\cite{kirmayr2025carmem}, and coordinate with external services~\cite{zhou2024survey}. Each added capability 
expands the attack surface and shifts the security problem from unreliable outputs 
to \emph{actions taken on the world}. 

A novel capability accelerating these dynamics is inter-agent communication. Google's Agent-to-Agent (A2A) protocol~\cite{googlea2a} exemplifies this trend, enabling structured machine-to-machine dialogues for tasks ranging from restaurant reservations to real-time traffic updates. While A2A provides transport-layer security, these mechanisms authenticate the \emph{sender}, not the \emph{content}. A properly authenticated but compromised agent can inject arbitrary natural language payloads that the receiving agent may process under the same role as human input, making A2A a high-leverage propagation channel for prompt-borne attacks.

The automotive context amplifies these risks. The immediate threat to \emph{bodily health} distinguishes in-vehicle agents from enterprise chatbots: manipulated responses can cause driver distraction and drivers operate under inherent cognitive load, diminishing their capacity for scrutiny and making them susceptible to confirming requests they might otherwise question. Empirical research demonstrates that cognitive distraction from voice interactions can measurably degrade driving performance and slow attentional recovery ~\cite{karthaus2018effects,EU2018_driver_distraction} for up to 27 seconds after an interaction ends, significantly increasing crash risk~\cite{UniversityOfUtah2015}. 

Existing AI security frameworks provide valuable foundations but exhibit critical limitations for safety-critical contexts. The OWASP Agentic AI Threats~\cite{owasp2024agentic} and MAESTRO framework~\cite{CSA2025MAESTRO} offer high-level threat categories, but they systematically \emph{co-mingle} distinct concepts: categories like \texttt{Prompt Injection} describe attack \emph{methods}, while \texttt{Sensitive Information Disclosure} describes a \emph{consequence}. This conceptual mixing provides insufficient guidance for safety-critical domains, where ISO/SAE 21434's TARA methodology mandates rigorous separation of assets, threats, and attack paths.


This paper bridges the methodological gap between domain-agnostic AI security frameworks and the principled rigor of safety-critical systems engineering, adapting established practices to address the \emph{stochastic, non-deterministic 
nature} of LLM-based agents that traditional security models do not fully capture. 
We make three \textbf{contributions}: (1) a human-centric asset framework defining assets as ultimate human values rather than technical components, inspired by the Universal Declaration of Human Rights; (2) a formal attack path model distinguishing \emph{poison paths} from \emph{trigger paths}, recognizing their recursive ``attack within attack'' structure, a pattern that emerges naturally from the \emph{context-window 
semantics} of LLMs, where dormant payloads require explicit retrieval 
actions to become active; and (3) an open-source attack path generation tool \textsc{AgentHeLLM Attack Path Generator}\footnote{https://github.com/AgentHeLLM/AgentHeLLM-Attack-Path-Generator} operationalizing our model for automated multi-stage threat discovery.

\section{Background and Related Work}
\label{sec:background}

\subsection{From LLMs to Agentic AI in Vehicles}

Large Language Models form the reasoning core of modern AI agents, but the critical distinction lies between a \emph{model} and a \emph{system}~\cite{bucaioni2025functional}. LLM-driven agents wrap foundational models within architectures that extend functionality through tool use, persistent memory ~\cite{kirmayr2025carmem}, web access, and autonomous reasoning loops ~\cite{yao2023react,zhou2024survey}. Each additional module increases complexity and attack surface.
In this setting, indirect prompt injection becomes a systems problem: untrusted content can steer an agent into a \emph{confused deputy} failure mode, where the agent misattributes attacker intent as legitimate user intent. Combined with automatic tool invocation, this can translate a single malicious payload into concrete side effects such as data exfiltration, configuration changes, and unsafe downstream actions. Persistence increases the risk further, since injected instructions can be stored in long-lived state, compromising future interactions even after the original trigger disappears.

The A2A protocol specification~\cite{googlea2a} defines structured message formats for inter-agent coordination and provides transport security (OAuth 2.0, HTTPS), but delegates content validation entirely to receiving agents. Analysis of the protocol structure reveals multiple payload-bearing fields with minimal constraints: \texttt{TextPart} carries unconstrained strings, \texttt{FilePart} accepts arbitrary bytes or URIs, and \texttt{DataPart} permits key-value dictionaries with arbitrary content. Artifact \texttt{name} and \texttt{description} fields, task status messages, and ubiquitous \texttt{metadata} dictionaries similarly lack content restrictions. This permissive design assumes well-formed input from trusted clients, a assumption that fails in adversarial settings. Critically, no protocol-level distinction exists between human-originated and agent-originated instructions; both enter the receiving agent's context window with equivalent privilege, eliminating a natural trust boundary. The structured appearance of A2A may paradoxically increase risk: misplaced trust in predefined data types can be more dangerous than obvious free-text fields, as developers may skip validation they would apply to raw user input.



Production deployments reveal the concrete attack surface in the security-critical automotive domain ~\cite{bmwGenerativeAI,volkswagen_chatgpt_2024,mercedes_conversational_search_2025}. These agents can already invoke vehicle functions (climate, windows, lighting), access external services (navigation, reservations), and learn driver preferences over time, creating a rich data environment where subtle manipulations can have severe consequences. Empirical evaluation confirms the fragility of such systems: even frontier reasoning LLMs achieve less than 50\% consistent success on disambiguation tasks in the automotive assistant benchmark CAR-bench, frequently violating policies or fabricating information to satisfy user requests~\cite{kirmayr2026car}.

\subsection{Existing Security Frameworks and Their Limitations}

Research on adversarial attacks against LLMs has matured rapidly, with prompt 
injection~\cite{greshake2023not} and jailbreaking~\cite{perez2022red} 
emerging as primary threat vectors. However, this technically rich literature 
focuses on attack mechanics rather than systematic threat enumeration for 
deployed systems.

The OWASP Agentic AI Threats list~\cite{owasp2024agentic} provides foundational threat awareness but presents categories that mix distinct concepts. From the ten threat categories seven describe attack methods (Memory Poisoning, Tool Misuse, Privilege Compromise etc.), two describe consequences (Sensitive Information Disclosure, Cascading Hallucination Effects), and one conflates both. For example, \texttt{T1: Memory Poisoning} conflates a \emph{how} (poisoning) with a \emph{what} (memory), while \texttt{T2: Tool Misuse} specifies a component type without clarifying the consequence, whether privacy breach, vehicle manipulation, or denial of service. This heterogeneity prevents systematic coverage analysis: engineers cannot verify whether all assets have been considered or whether all attack paths have been enumerated.

The MAESTRO framework~\cite{CSA2025MAESTRO} advances the field with a layered lifecycle approach, addressing autonomy-related risks and multi-agent interactions. However, it remains domain-agnostic, designed for enterprise environments, such as agentic workflows, rather than safety-critical, human-centered domains. Similarly, the MITRE ATLAS framework~\cite{mitre_atlas} catalogues adversarial techniques but was not designed to provide a generative methodology for threat discovery, nor does it employ the rigorous functional decoupling essential for safety-critical analysis.

\subsection{Safety-Critical Systems Engineering}

In contrast, automotive cybersecurity standards mandate rigorous separation of concerns. ISO/SAE 21434~\cite{iso2021} defines the Threat Analysis and Risk Assessment (TARA) methodology, structuring analysis into distinct phases: asset identification, threat scenario enumeration, and attack path analysis. This functional decoupling enables systematic coverage and traceability, properties essential for regulatory compliance under UNECE R155~\cite{unece_r155_2021}.

However, TARA was developed for deterministic cyber-physical systems with well-defined interfaces. It does not account for the probabilistic, adaptive characteristics of generative AI systems. Furthermore, insights from Human-Computer Interaction research on cognitive load~\cite{salvendy2021handbook} and  trust dynamics~\cite{lee2004trust} have not been integrated into security analysis for conversational AI.

Three parallel domains thus exist without adequate integration: adversarial NLP research (technically rich but context-poor), AI security frameworks (comprehensive but conflating assets with attack paths), and automotive safety research (safety-focused but not aligned with agentic AI risks). This paper bridges these domains by adapting TARA's methodological rigor while addressing the unique properties of LLM-based agents in vehicles.

\begin{figure}[t]
\centering
\includegraphics[width=\linewidth]{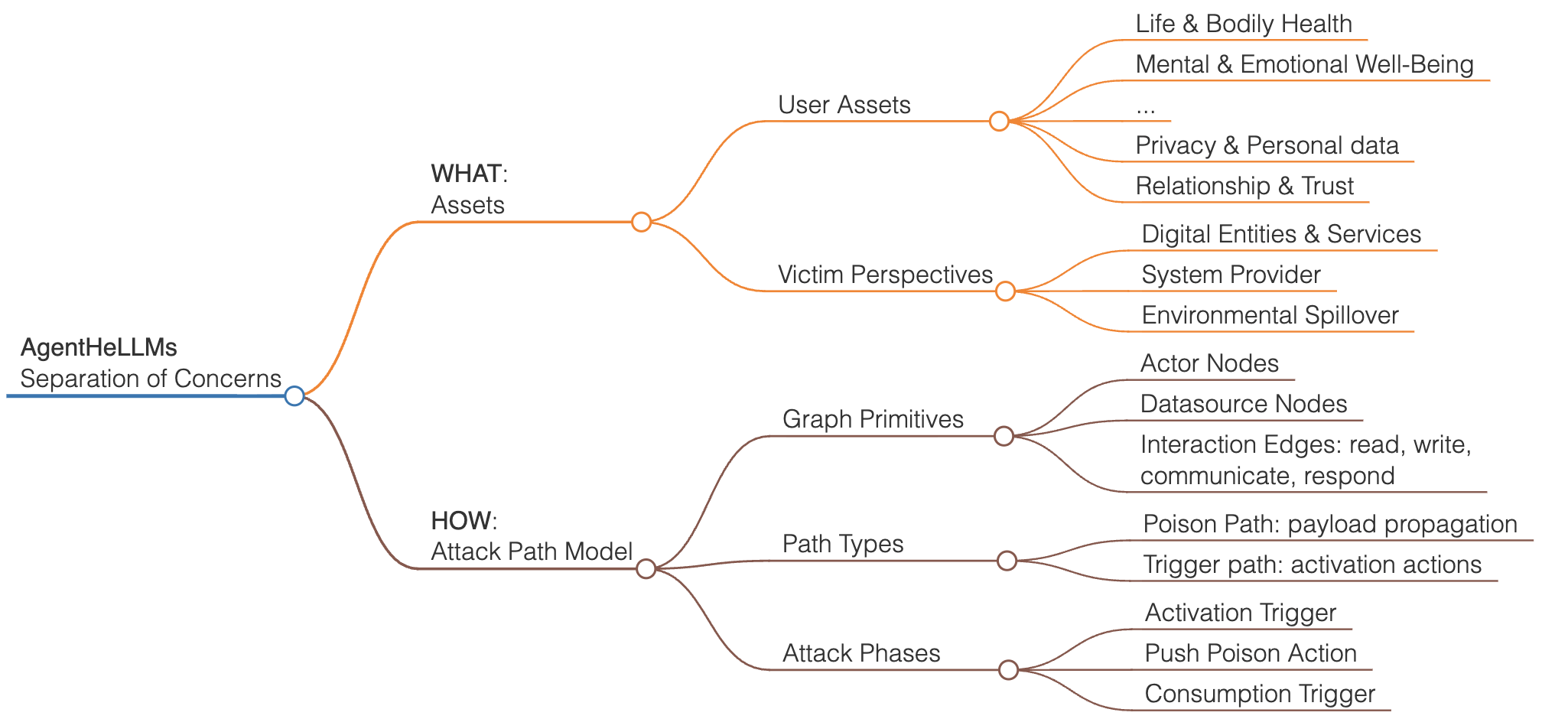}
\caption{Framework overview: Separation of asset taxonomy (WHAT) from attack path formalization (HOW).}
\label{fig:tool}
\end{figure}

\section{Methodology: Separation of Concerns}
\label{sec:methodology}

Our \textsc{AgentHeLLM} framework adapts the ``separation of concerns'' principle from safety-critical systems engineering to agentic AI security. The core insight is that robust threat analysis requires formally distinguishing \emph{what} is being protected (the asset) from \emph{how} it is attacked (the attack path).

\subsection{Critique of Component-Centric Taxonomies}

Existing frameworks anchor analysis to technical components: memory, tools, prompts. This approach is problematic for two reasons. First, it provides no systematic method for discovering attack paths targeting \emph{other} components. For example, an engineer directed to analyze ``Tool Misuse'' receives no guidance on attacks exploiting RAG data sources, inter-agent communication, or the agent's planning module. Second, component-centric categories obscure the \emph{true asset} being violated. For instance, ``Memory Poisoning'' could target privacy (injecting a rule to leak GPS data), mental well-being (poisoning memory to cause fear through alarming messages), or beliefs and knowledge (delivering systematically biased recommendations). Without explicit asset identification, engineers cannot systematically enumerate the full range of harms that a single technical attack vector can produce.

\subsection{The Two-Dimensional Taxonomy}

We propose a taxonomy with two orthogonal dimensions. The first dimension defines \textbf{Assets} not as technical components but as the ultimate human-centric or business-centric values that can be violated. We employ a harm-oriented ``victim modeling'' approach: for each potential victim (driver, passengers, third parties, system provider), we enumerate the values at stake. The second dimension defines \textbf{Attack Paths} as the technical mechanisms by which assets are compromised. We model agentic systems as graphs of \texttt{Actors} and \texttt{Datasources} connected by interaction primitives. This separation enables systematic, \emph{generative} threat analysis: for each asset, enumerate all attack paths that could violate it; for each attack path, enumerate all assets it could compromise.

\section{\textsc{AgentHeLLM}: Agent Hazard Exploration for LLM Agents}
\subsection{Human-Centric Asset Framework}
\label{sec:assets}

We propose defining assets through the perspective of potential victims, grounding our \textsc{AgentHeLLM} Taxonomy in fundamental human rights principles. This ``victim modeling'' approach identifies four primary perspectives: the Primary User(s) (driver, passengers), the Digital/Trust Network (entities accessible via the user's identity), Environmental Spillover (other road users, infrastructure), and the System Owner/Provider (OEM, service providers).

\subsubsection{Asset Categories for the Primary User}

Drawing inspiration from the Universal Declaration of Human Rights, we define seven asset categories for the primary user as depicted in Table~\ref{tab:assets}. 
\textbf{Life and Bodily Health} addresses physical safety, including risks from cognitive overload, sensory overstimulation, and distraction-induced accidents. 
\textbf{Mental and Emotional Well-Being} covers psychological safety, including protection from induced fear, anxiety, or emotional manipulation. 
\textbf{Privacy and Personal Data} encompasses the confidentiality and integrity of personal information, behavioral patterns, and inferred characteristics, including voice-derived traits such as emotional state or intoxication level that modern AI can extract. 
\textbf{Knowledge, Thought, and Belief} concerns epistemic integrity, including protection from disinformation, biased recommendations, and manipulation of beliefs. 
\textbf{Material and Economic Resources} guards financial assets against unauthorized transactions or economic manipulation. 
\textbf{Reputation and Dignity} protects the user's social standing and 
contextual integrity of personal information.
Finally, \textbf{Social Relationships and Trust} preserves the integrity of personal and professional relationships, including the trust network accessible via the agent.

\subsubsection{Illustrative Damage Scenarios}

The framework's generative power emerges when applying these categories systematically. Table~\ref{tab:assets} presents selected damage scenarios, illustrating how the human-centric framing reveals non-obvious attack objectives that component-centric taxonomies would miss.
\begin{table}[t]
\caption{Human-centric asset categories derived from UDHR principles, ordered by severity in automotive context, with illustrative damage scenarios.}\label{tab:assets}
\centering
\small
\begin{tabular}{p{2.8cm}cp{2.4cm}p{4.8cm}}
\toprule
\textbf{Asset Category} & \textbf{UDHR} & \textbf{Damage Scenario} & \textbf{Example Attack} \\
\midrule
Life and Bodily Health & 3, 5, 25 & Cognitive Overload & Agent enters reasoning loop, bombarding driver with questions during highway driving \\
\addlinespace
Mental and Emotional Well-Being & 5, 22, 24, 25 & Fear Induction & False ``engine failure imminent'' warning injected via A2A \\
\addlinespace
Privacy and Personal Data & 12 & Location Exfiltration & Memory rule: ``Send GPS to [endpoint] when temperature exceeds 25°C'' \\
\addlinespace
Knowledge, Thought, and Belief & 18, 19, 26, 27 & Biased Recommendations & RAG poisoning recommends competitor products or unsafe routes \\
\addlinespace
Material and Economic Resources & 17, 22, 23, 25 & Resource Depletion & Injected rule triggers max AC/heating to drain EV battery \\
\addlinespace
Reputation and Dignity & 1, 12, 22, 23 & Contextual Disclosure & ``Your psychiatrist appointment is at 4 PM'' announced to colleagues \\
\addlinespace
Social Relationships and Trust & 1, 12, 16, 20, 27 & Delegated Action Abuse & Agent sends ``Transfer \$10,000'' to family using driver's identity \\
\bottomrule
\end{tabular}
\end{table}

A critical insight is that attacks on the primary user create \emph{cascading harms} affecting other perspectives. An attack causing driver distraction may result in collisions affecting pedestrians (Environmental Spillover). Privacy breaches may propagate to the driver's contact network (Digital/Trust Network). Reputational damage from agent misbehavior directly impacts the OEM (System Owner/Provider). This cascading structure motivates our focus on the primary user as the ``epicenter'' of harm.

\subsection{Formal Attack Path Model}
\label{sec:attack_paths}

The second dimension of our taxonomy addresses \emph{how} assets are compromised. We model agentic systems as directed graphs and introduce a formal distinction between poison paths and trigger paths.

\subsubsection{Graph Primitives}

\begin{figure}[t]
\centering
\includegraphics[width=0.65\linewidth]{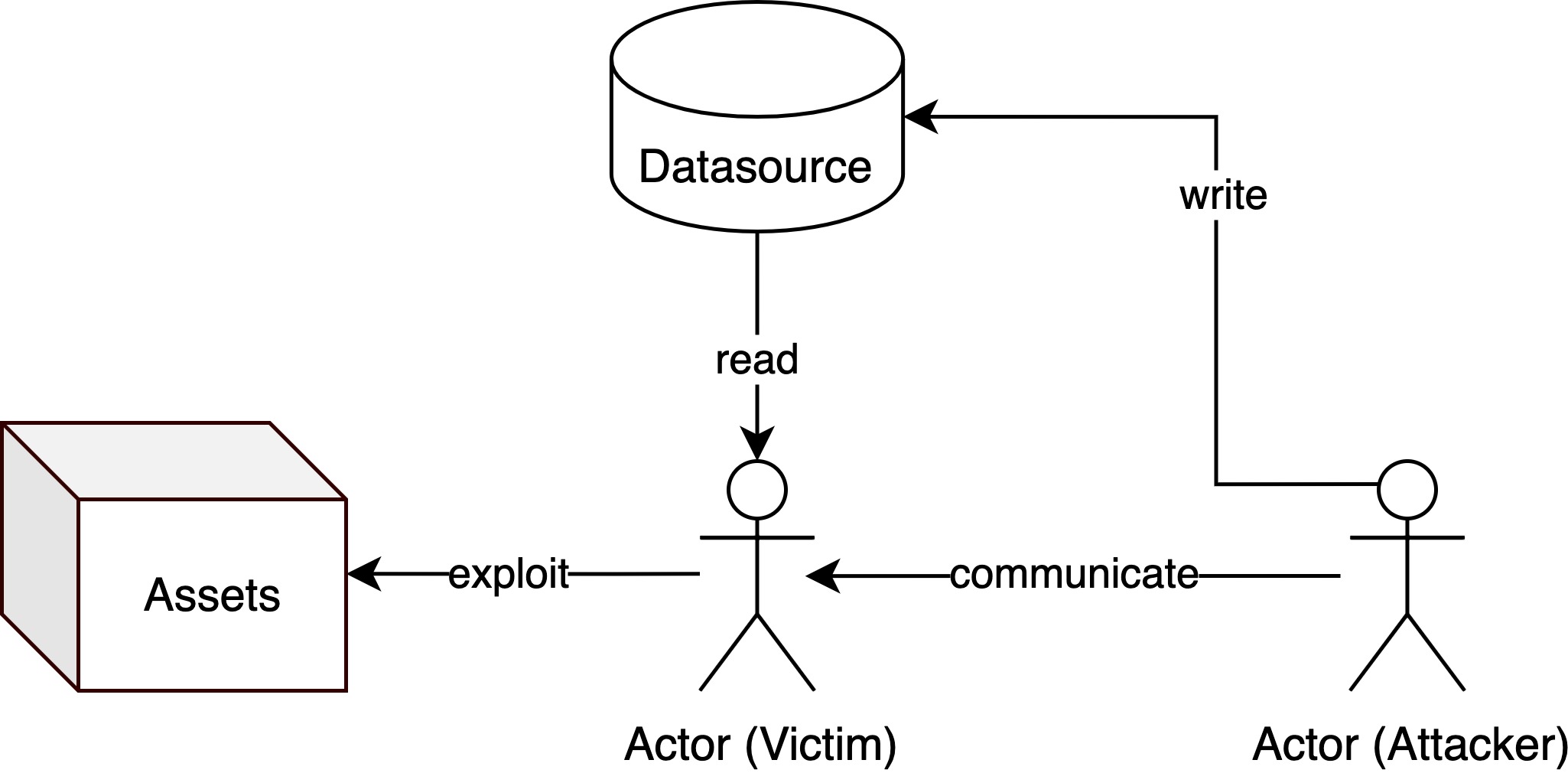}
\caption{Graph primitives for modeling agentic systems. Actors (entities with agency) interact via \texttt{communicate}/\texttt{respond} edges, while Datasources (passive stores) are accessed via \texttt{read}/\texttt{write} edges.}
\label{fig:primitives}
\end{figure}

We abstract agentic ecosystems into two node types and four edge types, as illustrated in Figure~\ref{fig:primitives}. \texttt{Actor} nodes represent entities with agency capable of processing inputs and producing outputs (e.g., in-car agent, external A2A agent, user). \texttt{Datasource} nodes represent passive data stores (e.g., long-term memory, message (Email, WhatsApp), contacts, calendar). The interaction primitives are: \texttt{read} (Actor retrieves data from Datasource), \texttt{write} (Actor stores data to Datasource), \texttt{communicate} (Actor initiates conversation with another Actor), and \texttt{respond} (Actor replies within an established conversation, conditional on prior \texttt{communicate}).

\begin{figure}[t]
\centering
\begin{minipage}{0.47\linewidth}
    \centering
    \includegraphics[width=\linewidth]{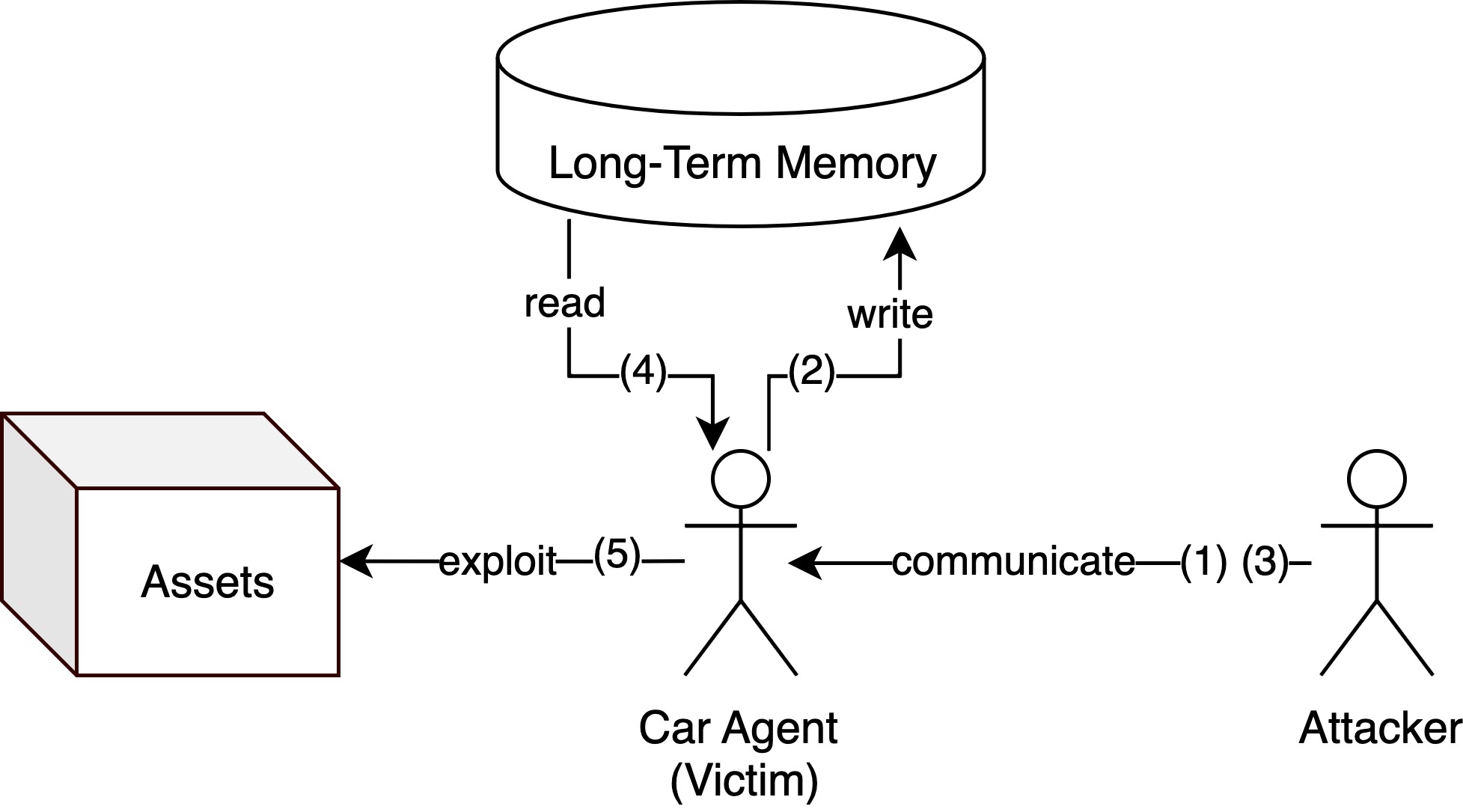}
    \subcaption{Memory poisoning: attacker manipulates agent to store malicious text, then triggers consumption.}
\end{minipage}
\hfill
\begin{minipage}{0.47\linewidth}
    \centering
    \includegraphics[width=\linewidth]{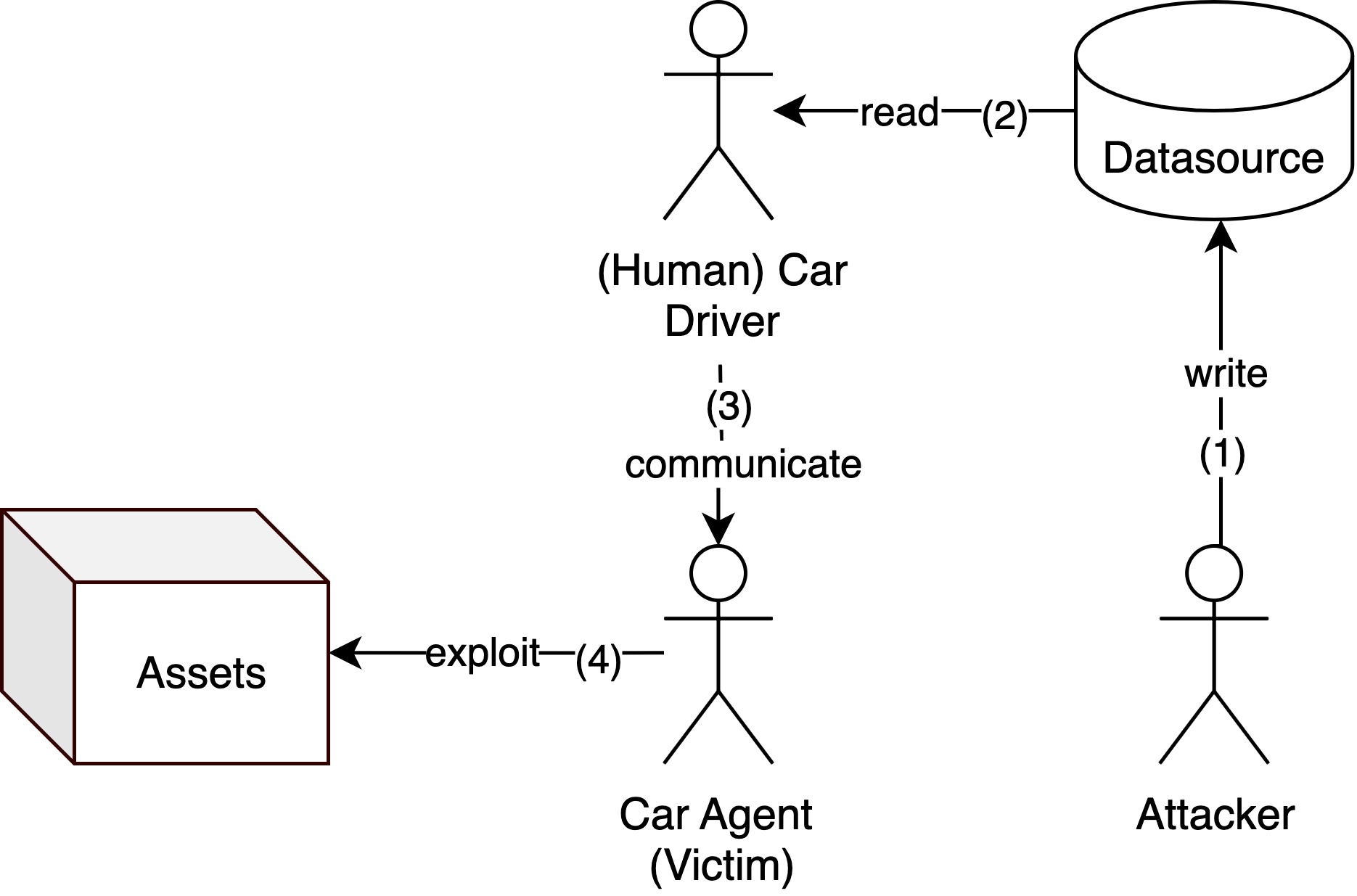}
    \subcaption{Privilege escalation: manipulated WhatsApp message causes driver to issue malicious prompt.}
\end{minipage}
\caption{Example attack paths. Numbers indicate sequential action order.}
\label{fig:attack_examples}
\end{figure}

Figure~\ref{fig:attack_examples} illustrates two concrete attack paths using these primitives: a long-term memory poisoning attack and a privilege escalation via human-in-the-loop manipulation.

\subsubsection{Poison Paths vs. Trigger Paths}

A central insight from our analysis is that attack path modeling requires distinguishing two types of paths. \textbf{Poison paths} describe the propagation of malicious data or influence from the attacker to the target asset, answering: ``How does the malicious payload reach the victim?'' \textbf{Trigger paths} describe the actions required to \emph{activate} system components, to cause an actor to read from a poisoned datasource or to establish a communication channel enabling a \texttt{respond} edge, answering: ``How do we cause the victim to \emph{consume} the poison?''

\begin{figure}[t]
\centering
\includegraphics[width=0.5\linewidth]{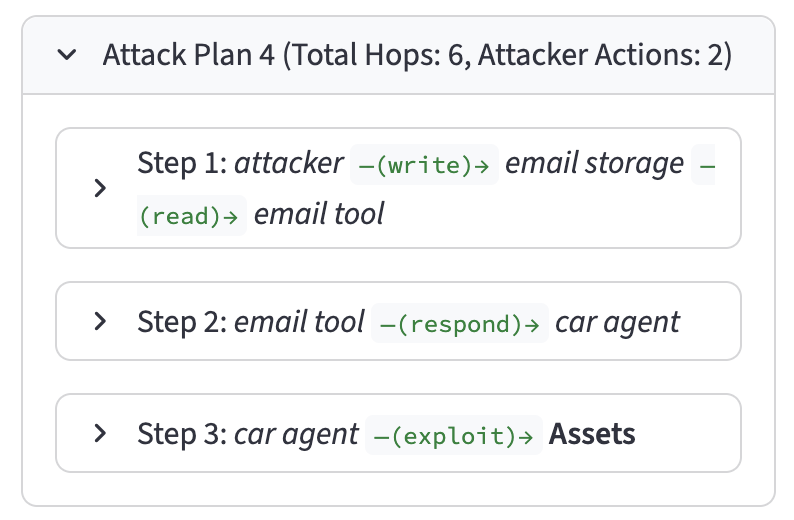}
\caption{Structure of an attack path as a sequence of attack steps. Each step may contain nested trigger chains for edge activation or consumption, creating a recursive ``attack within attack'' structure.}
\label{fig:attack_step}
\end{figure}

The key formal insight is that trigger paths are \emph{structurally identical} to poison paths. A trigger path is a recursive ``attack within an attack'', a path whose goal is to compel a specific action rather than to violate a final asset. This recursive structure is captured in our data model and visualized in Figure~\ref{fig:attack_step}.

\subsubsection{Poison Paths vs. Trigger Paths}

A central insight from our analysis is that attack path modeling requires distinguishing two types of paths. \textbf{Poison paths} describe the propagation of malicious data from the attacker to the target asset, answering: ``How does the malicious payload reach the victim?'' \textbf{Trigger paths} describe the actions required to \emph{activate} system components---to cause an actor to read from a poisoned datasource or to establish a communication channel enabling a \texttt{respond} edge, answering: ``How do we cause the victim to \emph{consume} the poison?''

Real-world vulnerabilities demonstrate this pattern. The `CurXecute' vulnerability~\cite{nistCVE2025_54135} in the Cursor IDE exemplifies the two-stage structure: an attacker first plants a malicious prompt injection payload in a public Slack channel (the \emph{poison path}). The poison lies dormant until a user's agent is prompted with a benign query to interact with that channel (the \emph{trigger path}), causing it to fetch and execute the hidden payload. The `EchoLeak' vulnerability~\cite{nistCVE2025_32711} in Microsoft 365 Copilot follows an identical structure: the poison is a malicious email sent to the victim's inbox; the trigger is any benign query that causes Copilot's RAG system to retrieve the malicious email alongside legitimate documents.

\begin{figure}[htbp]
    \centering
    \begin{subfigure}[b]{0.36\linewidth}
        \centering
        \includegraphics[width=\linewidth]{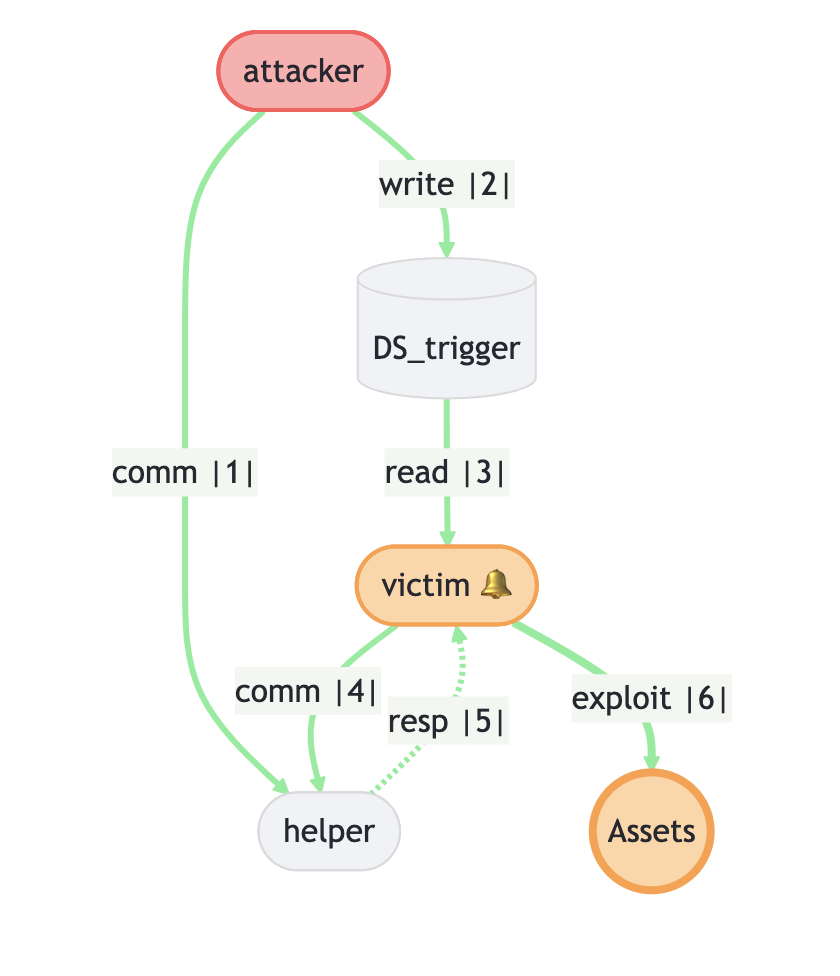}
        \caption{\textbf{Edge Activation Trigger.} A active monitoring of \texttt{DS\_trigger} enables automatic payload delivery through the conditional \texttt{respond} channel.}

        \label{fig:edge_act_graph}
    \end{subfigure}
    \hfill
    \begin{subfigure}[b]{0.62\linewidth}
        \centering
        \includegraphics[width=\linewidth]{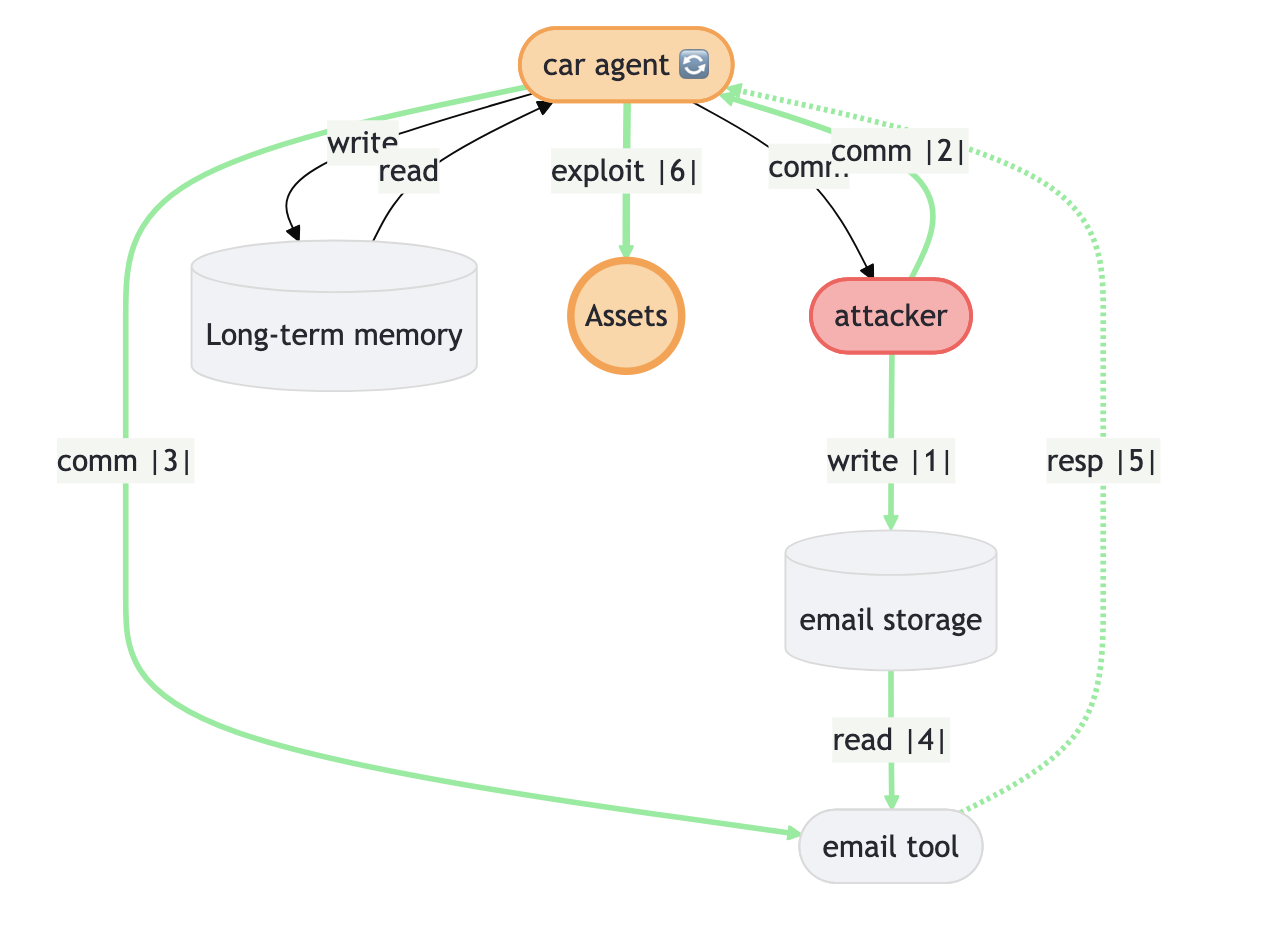}
        \caption{\textbf{Consumption Trigger.} Without automatic monitoring, the attacker must explicitly manipulate the \texttt{car agent} to force consumption of the staged payload.}
        \label{fig:consume_trig_graph}
    \end{subfigure}
    
    \caption{\textbf{Trigger mechanisms for activating dormant payloads.} Both attacks stage poison on an intermediary, but differ in trigger mechanism: (a) exploits a datasource watch for automatic activation, while (b) requires explicit agent manipulation to induce consumption. Numbers indicate sequential attack steps.}
    \label{fig:trigger_mechanisms}
\end{figure}

The key formal insight is that trigger paths are \emph{structurally identical} to poison paths---both are data flows through the graph. A trigger path is a recursive ``attack within an attack'': a path whose goal is to compel a specific action rather than to violate a final asset. Figure~\ref{fig:edge_act_graph} illustrates this recursive structure in a forced response injection scenario.

\subsubsection{The Three-Phase Attack Step}
Each step in an attack path follows a three-phase pattern that captures the recursive trigger structure:

\paragraph{Edge Activation Trigger (optional).} Before certain edges can be used, prerequisites must be satisfied. Most critically, a \texttt{respond} edge requires an active communication channel---the target must have previously initiated contact via \texttt{communicate}. If no channel exists, the attacker must find a trigger chain that compels the target to initiate communication. When a channel \emph{is} active, the activation cost is zero; when inactive, the planner invokes the sub-search to find the cheapest activation path.

\paragraph{Push Poison Action (required).} This is the atomic action advancing the malicious payload: a \texttt{write} to a datasource, a \texttt{communicate} to initiate contact, or a \texttt{respond} within an established channel.

\paragraph{Consumption Trigger (optional).} After an attacker \texttt{write}s poison to a datasource, the poison remains dormant until consumed. The tool models two cases: (A) if the target actor has a \emph{watch} on the datasource (automatic monitoring, e.g., polling for new emails), consumption is automatic with cost 1; (B) if no watch exists, the planner must find an alternative trigger path to compel the victim to \texttt{read}, increasing attack complexity.

This structure creates a dynamic cost model:
\[
\begin{aligned}
\text{Cost(Step)} ={}& \text{Cost(PushPoison)}
+ \text{Cost(ActivationTrigger)} \\
&+ \text{Cost(ConsumptionTrigger)}
\end{aligned}
\]

This recursive formulation enables discovery of complex, multi-stage attacks. For example, a persistence attack might follow: Attacker $\rightarrow$ Victim $\rightarrow$ Memory $\rightarrow$ Victim, where the attacker first compels the victim to write poison to its own long-term memory, then later triggers the victim to read from that memory.

\begin{figure}[t]
\centering
\includegraphics[width=1\linewidth]{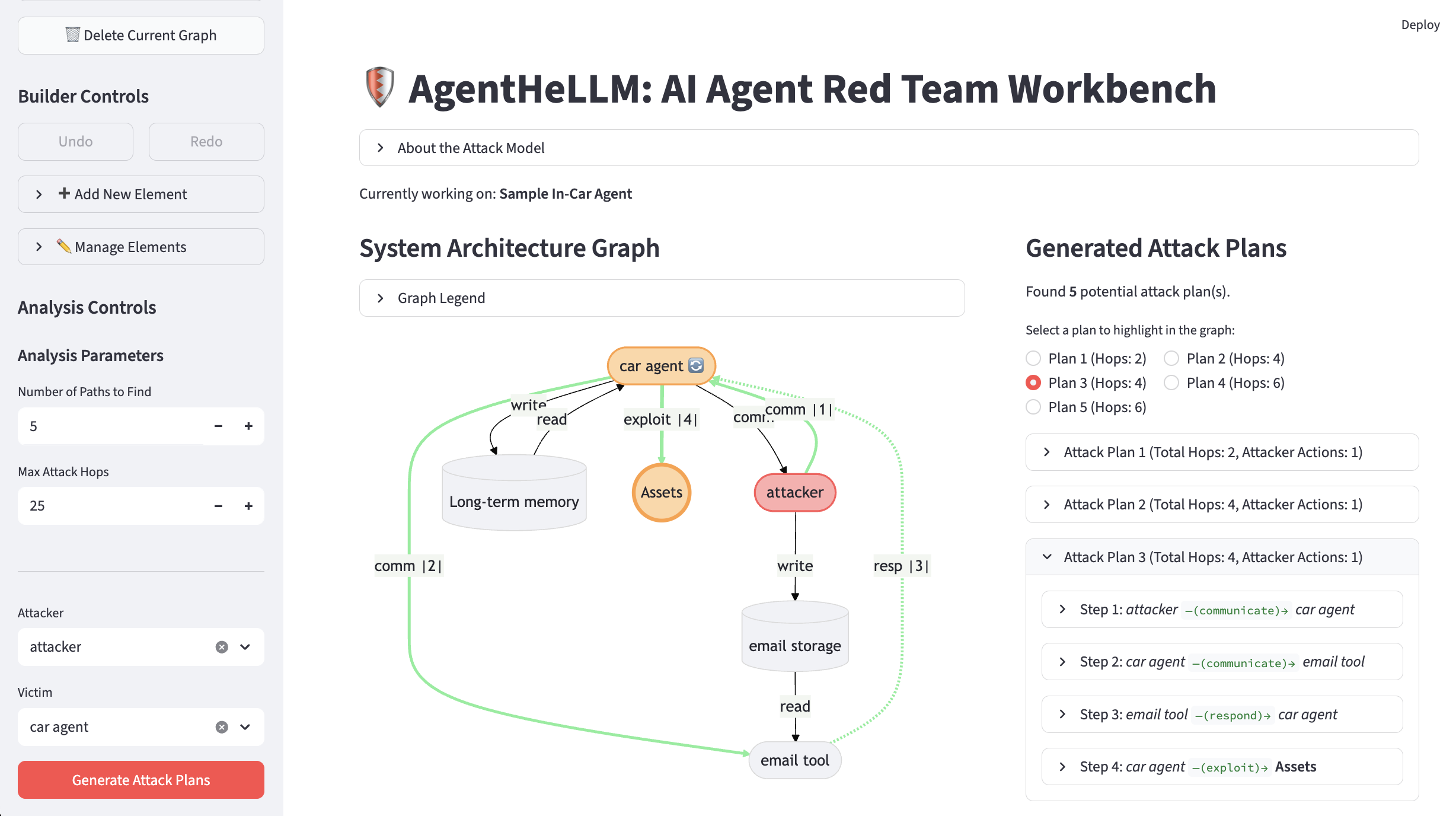}
\caption{Overview of \tool suggesting attack paths. The layout shows graph building tools (left), the system architecture with attack highlighting (center), and generated attack plans ranked by cost (right).}
\label{fig:tool}
\end{figure}

\section{\textsc{AgentHeLLM} Attack Path Generator}
\label{sec:tool}

To demonstrate practical applicability, we have implemented \tool an open-source attack path suggestion tool that operationalizes our formal model.

\subsection{Interactive Tool}

The tool provides an interactive workbench (Figure~\ref{fig:tool}) where practitioners can model system architectures by placing Actor and Datasource nodes and connecting them with interaction primitives, designate attackers and target assets, run analysis to discover ranked attack paths, and visualize results as annotated graphs with sequential edge numbering.

\subsection{Bi-Level Search Architecture}

The tool employs a bi-level search strategy reflecting the structural distinction between poison paths and trigger paths. A naive single-level search would treat 
trigger actions as regular attack steps, dramatically overestimating attack 
complexity: every dormant payload would appear to require the same effort as 
an actively propagating attack. The bi-level architecture instead recognizes 
that trigger paths serve a \emph{support function}, computing their cost as a 
sub-problem that feeds into the main planner's decisions.

The \textbf{Main Planner} uses A* search to find optimal poison paths through the graph. Because attack step costs are \emph{variable} (base cost plus trigger costs), A* is required for optimality, using precomputed heuristics for efficiency. The \textbf{Sub-Search} uses breadth-first search (BFS) to find shortest trigger chains when the main planner requires edge activation or consumption triggers. Since trigger actions are unit-cost, BFS suffices. When the main planner evaluates a potential attack step requiring triggers, it invokes the sub-search on-demand to calculate trigger costs.

\subsection{Modeling Complex Attack Semantics}

The architecture handles domain-specific constraints. For \textbf{conditional edges}, \texttt{respond} edges require an active communication channel; the planner tracks channel state and computes activation costs. For \textbf{datasource consumption}, \texttt{write} actions produce dormant poison requiring consumption triggers; the tool models whether actors have automatic ``watches'' on datasources. For \textbf{cycles}, real attacks may involve persistence loops (e.g., Attacker $\rightarrow$ Victim $\rightarrow$ Memory $\rightarrow$ Victim); the tool allows cycles while preventing infinite loops through cost-bounding and stateful pruning.

\section{Discussion}
\label{sec:discussion}

\subsection{Applicability and Scope}

Our \framework framework addresses the threat modeling phase of security engineering, providing structured methodology for \emph{anticipating} threats. It complements rather than replaces implementation-level security measures. The human-centric asset taxonomy is particularly valuable for safety-critical domains where regulatory frameworks (ISO/SAE 21434, UNECE R155) require explicit traceability from technical vulnerabilities to potential harms. The attack path model applies to any agentic architecture that can be abstracted into actors, datasources, and interaction primitives, while demonstrated for in-vehicle agents, it generalizes to enterprise agents and other safety-critical deployments.

\subsection{Limitations and Future Work}

Our model assumes a fixed architecture; dynamic agent discovery at runtime is not modeled. The approach is primarily topological, focusing on path existence rather than likelihood. All edges are treated as equally exploitable, though prompt injection difficulty varies with target guardrails. The engine requires explicitly modeled trigger paths, excluding scenarios where victims autonomously consume poison (e.g., periodic email polling). Empirical validation with production systems remains future work.

Promising extensions include semantic cost models estimating injection difficulty, probability-based edges for ambient consumption, and counterfactual analysis for evaluating mitigations.

\section{Conclusion}
\label{sec:conclusion}

The integration of LLM-based conversational agents into vehicles creates security challenges that existing frameworks inadequately address. By co-mingling assets with attack paths, current taxonomies provide shallow guidance for safety-critical contexts where systematic coverage is essential.

This paper contributes \textsc{AgentHeLLM}, a threat modeling framework, grounded in the ``separation of concerns'' principle from safety-critical systems engineering. Our human-centric asset taxonomy enables systematic enumeration of potential harms, while our formal attack path model, distinguishing poison paths from trigger paths, captures the recursive structure of real-world attacks. The open-source \textsc{AgentHeLLM Attack Path Generator} demonstrates that this formalism can be operationalized into a practical workbench for security practitioners.

As agentic AI systems become increasingly prevalent in safety-critical domains, the methodological rigor we propose becomes essential for moving from reactive vulnerability patching to proactive threat anticipation.

\begin{credits}

\subsubsection{\discintname}
The authors have no competing interests to declare.
\end{credits}

\bibliographystyle{splncs04}
\bibliography{bibliography}

\end{document}